\documentclass[10pt,twocolumn,letterpaper]{article}

\usepackage{cvpr}              
\definecolor{cvprblue}{rgb}{0.21,0.49,0.74}
\usepackage[pagebackref,breaklinks,colorlinks,allcolors=cvprblue]{hyperref}

\usepackage{multirow}
\usepackage{bm}
\usepackage{xcolor}
\usepackage{colortbl}
\usepackage[fixed]{fontawesome5}
\usepackage[accsupp]{axessibility} 

\def\paperID{} 
\def\confName{CVPR}
\def\confYear{2026}

\title{Tell Model Where to Look: Mitigating Hallucinations in MLLMs by Vision-Guided Attention}

\author{
    Jianfei Zhao\textsuperscript{\rm 1,2},
    Feng Zhang\textsuperscript{\rm 1},
    Xin Sun\textsuperscript{\rm 1,}\thanks{Corresponding Authors.},
    Chong Feng\textsuperscript{\rm 1,3,}\footnotemark[1],
    Zhixing Tan\textsuperscript{\rm 4,}\footnotemark[1]\\
    \textsuperscript{\rm 1}School of Computer Science and Technology, Beijing Institute of Technology\\
    \textsuperscript{\rm 2}Zhongguancun Academy\\
    \textsuperscript{\rm 3}Southeast Academy of Information Technology, Beijing Institute of Technology\\
    \textsuperscript{\rm 4}Zhongguancun Laboratory\\
     {\tt\small \{ zhqingan, bit\_zhangfeng, sunxin, fengchong\}@bit.edu.cn, zxtan@zgclab.edu.cn}
}

\begin{document}
\maketitle
\begin{abstract}
Visual attention serves as the primary mechanism through which MLLMs interpret visual information; however, its limited localization capability often leads to hallucinations. We observe that although MLLMs can accurately extract visual semantics from visual tokens, they fail to fully leverage this advantage during subsequent inference.
To address this limitation, we propose \textbf{V}ision-\textbf{G}uided \textbf{A}ttention (VGA), a training-free method that first constructs precise visual grounding by exploiting the semantic content of visual tokens, and then uses this grounding to guide the model's focus toward relevant visual regions. In image captioning, VGA further refines this guidance dynamically during generation by suppressing regions that have already been described.
In VGA, each token undergoes only a single forward pass, introducing a negligible latency overhead. In addition, VGA is fully compatible with efficient attention implementations such as FlashAttention.
Extensive experiments across diverse MLLMs and multiple hallucination benchmarks demonstrate that VGA achieves state-of-the-art dehallucination performance. Further analysis confirms that explicit visual guidance plays a crucial role in enhancing the visual understanding capabilities of MLLMs.
\footnote{The code is available at: \url{https://github.com/beta-nlp/VGA}.}

\end{abstract}    
\section{Introduction}
\label{sec:intro}

Multimodal Large Language Models (MLLMs) extend the capabilities of large language models (LLMs) into the visual domain, enabling tasks such as visual question answering and image captioning. However, their visual understanding remains limited, as they often exhibit hallucinations—producing outputs that contradict the actual visual content. This hallucination issue significantly affects the usability and reliability of MLLMs.

\begin{figure}
    \centering
    \includegraphics[width=0.8\linewidth]{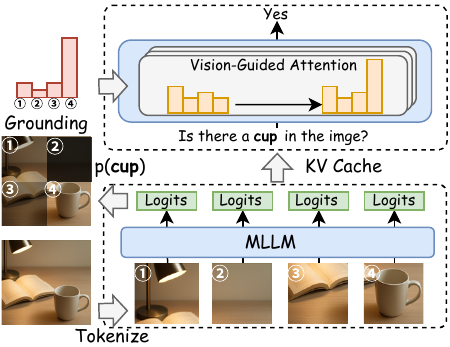}
    \caption{
    A diagram of vision‑guided attention. We first leverage the semantic features embedded in visual tokens to establish visual grounding, and then guide the model's visual attention toward the relevant image regions.
    }
    \label{fig:diagram}
\end{figure}

Various approaches have been proposed to mitigate hallucinations, such as constructing targeted datasets~\cite{DPA} or designing specialized loss functions~\cite{HSA-DPO,SymMPO}. However, the rapid iteration of model architectures has led to diminishing returns for training-based methods. In contrast, training-free hallucination mitigation techniques offer greater practical value due to their plug-and-play nature and strong generalization capability. In particular, a line of research focuses on optimizing visual attention—a mechanism that serves as the primary conduit through which multimodal large language models (MLLMs) interpret visual information~\cite{CInFlow,VInFlow}—providing one of the most direct pathways to reducing hallucinations.

Previous studies~\cite{PAI,VAF,CLVS,TARAC} have underscored the essential role of visual attention in MLLMs and shown that strengthening visual attention can substantially mitigate hallucination. However, these approaches depend on the inherent quality of visual attention itself, which is fundamentally limited in its ability to accurately localize critical visual regions~\cite{opera,VAR}, ultimately resulting in suboptimal performance.
While employing external tools~\cite{MARINE} or performing additional forward passes~\cite{AdaptVis} is feasible, these approaches introduce extra computational overhead and reduce practical usability. Furthermore, because they rely on optimization based on attention distributions, they are incompatible with FlashAttention~\cite{Flashattention}, which does not expose explicit attention weights, thereby restricting their applicability. Consequently, optimizing visual attention remains a significant challenge.

To our surprise, we found that the model can accurately extract semantic features from visual tokens and instantiate them as conditional probabilities within the visual logits. This probability distribution over the image enables token-level visual grounding. We term this mechanism Visual Semantic Confidence (VSC).
Based on this observation, we propose Vision-Guided Attention (VGA), a novel strategy illustrated in \Cref{fig:diagram} that leverages VSC-derived grounding to identify the most informative visual tokens and direct greater attention toward them.
More importantly, VGA does not require computing attention weights for individual tokens, making it fully compatible with FlashAttention.

In short, our main contributions are:
\begin{itemize}
    \item We introduce Visual Semantic Confidence (VSC), which leverages the semantic features of visual tokens to achieve precise visual grounding.
    \item We propose Vision-Guided Attention (VGA), a training-free and FlashAttention-compatible method that uses the visual grounding produced by VSC to guide visual attention, enabling the MLLM to focus on the most relevant visual tokens.
    \item Experiments and analyses demonstrate that VSC provides accurate visual grounding and that VGA effectively mitigates hallucinations in MLLMs.
    
\end{itemize}
\section{Related Work}
\label{sec:related work}

Constructing high-quality datasets~\cite{PerturboLLaVA, DPA} or designing hallucination-targeted optimization objectives~\cite{HSA-DPO, SymMPO} can directly improve model reliability, but the substantial training cost makes these approaches less cost‑effective. As a result, training‑free dehallucination methods have gained increasing attention. Integrating well‑established external tools can effectively enhance the reliability of MLLMs~\cite{Woodpecker, MARINE}, but this also increases the overall system cost. Consequently, optimizing the model's internal representations has become a more appealing alternative. Such methods mitigate hallucinations by adjusting internal components—such as attention weights, logits, or hidden states—to suppress hallucinatory signals.
A line of contrastive decoding–based approaches~\cite{CICD, VCD, DeGF, VISTA} directly reduces hallucinations in MLLMs by attenuating hallucinatory features in the logits. However, these methods generally require an additional forward pass to activate or expose such features. Other studies~\cite{DAMO, LookTwice, NDE} edit semantic features in the embedding space to enhance vision‑related representations; yet the high dimensionality and complexity of this space make such vector manipulations prone to introducing noise.

Attention mechanisms serve as the primary means by which MLLMs interpret visual features, and the distribution of attention weights offers an intuitive, interpretable reflection of the model's visual behavior.  
Several studies~\cite{PAI,IBD} have directly increased the model's attention to visual content, thereby empirically validating the critical role of visual attention in MLLMs.  
Other works~\cite{AD-HH, VHD, SEVI} have investigated functional heterogeneity among attention heads and sought to coordinate their roles accordingly.
Additionally, some approaches~\cite{VAR,VAF} attribute hallucinations to anomalous attention patterns and propose targeted refinements to address them.
While these methods effectively mitigate hallucinations by optimizing visual attention, they largely overlook the inherent variability among visual tokens—specifically, which tokens are most worthy of the model's attention.
Some approaches~\cite{CLVS, TARAC} leverage historical attention distributions to assess the importance of visual tokens and use this information to refine the current attention distribution. However, these methods rely entirely on the model's inherent visual behavior, which exhibits significant limitations.

Unlike previous works that exploit the limited potential of visual attention itself, we introduce a novel visual prior for visual attention. This prior is derived internally from the model and operates without reliance on external signals.
Specifically, we first leverage the semantic features of visual tokens to locate informative ones, and then construct visual guidance based on these locations to modulate the model's attention.

\section{Visual Semantic Confidence}
\begin{figure}
    \centering
    \includegraphics[width=\linewidth]{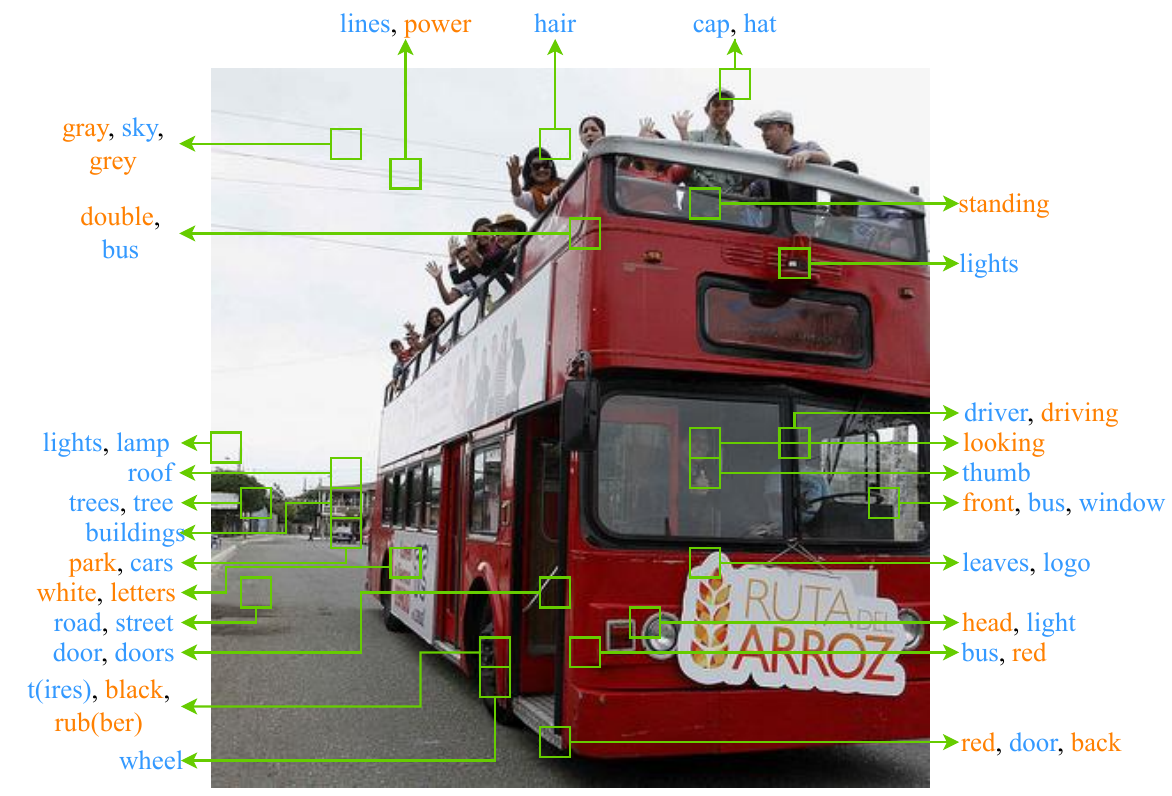}
    \caption{Top-3 tokens (excluding punctuation) in visual logits from LLaVA-1.5-7B.}
    \label{fig:vl_case}
\end{figure}

In MLLMs, an input image is first divided into patches, producing visual tokens $v_1, \cdots, v_m$.
After encoding by the visual encoder and projector, these visual tokens are concatenated with text embeddings to form a single sequence, which is then fed into the base LLM.
The final hidden states are passed through an unembedding matrix to generate logits over the vocabulary. Applying a softmax to the logit at the last position yields the conditional probability for the current generation context, which the model uses to predict the next token:
$p(t_{i}\mid t_{<i}) \propto \mathrm{softmax}(\mathrm{logit}_{t-1})$.

Logits provide a concrete representation of semantic information~\cite{IC}, where the score associated with each token reflects the model's confidence in the semantic consistency between that token and the current context. Analogously, visual logits capture the underlying visual semantics of each visual token; specifically, the score of a word within the visual logits quantifies the semantic correlation between that word and the corresponding visual context, as illustrated in \Cref{fig:vl_case}. We term this phenomenon in MLLMs Visual Semantic Confidence (VSC). By aggregating these semantic confidence scores across all visual tokens for a given word, we derive the word's overall semantic confidence relative to the entire image. Intuitively, \textbf{the VSC mechanism can enable token-level visual grounding}.
We discuss the VSC mechanism in two scenarios: \Cref{sec:object-directed} details visual grounding with direct visual demands (\emph{e.g.}, VQA), while \Cref{sec:object-agnostic} addresses cases without explicit visual demands (\emph{e.g.}, image captioning).

\subsection{Object-Directed Grounding}\label{sec:object-directed}
\begin{figure}
    \centering
    \includegraphics[width=0.7\linewidth]{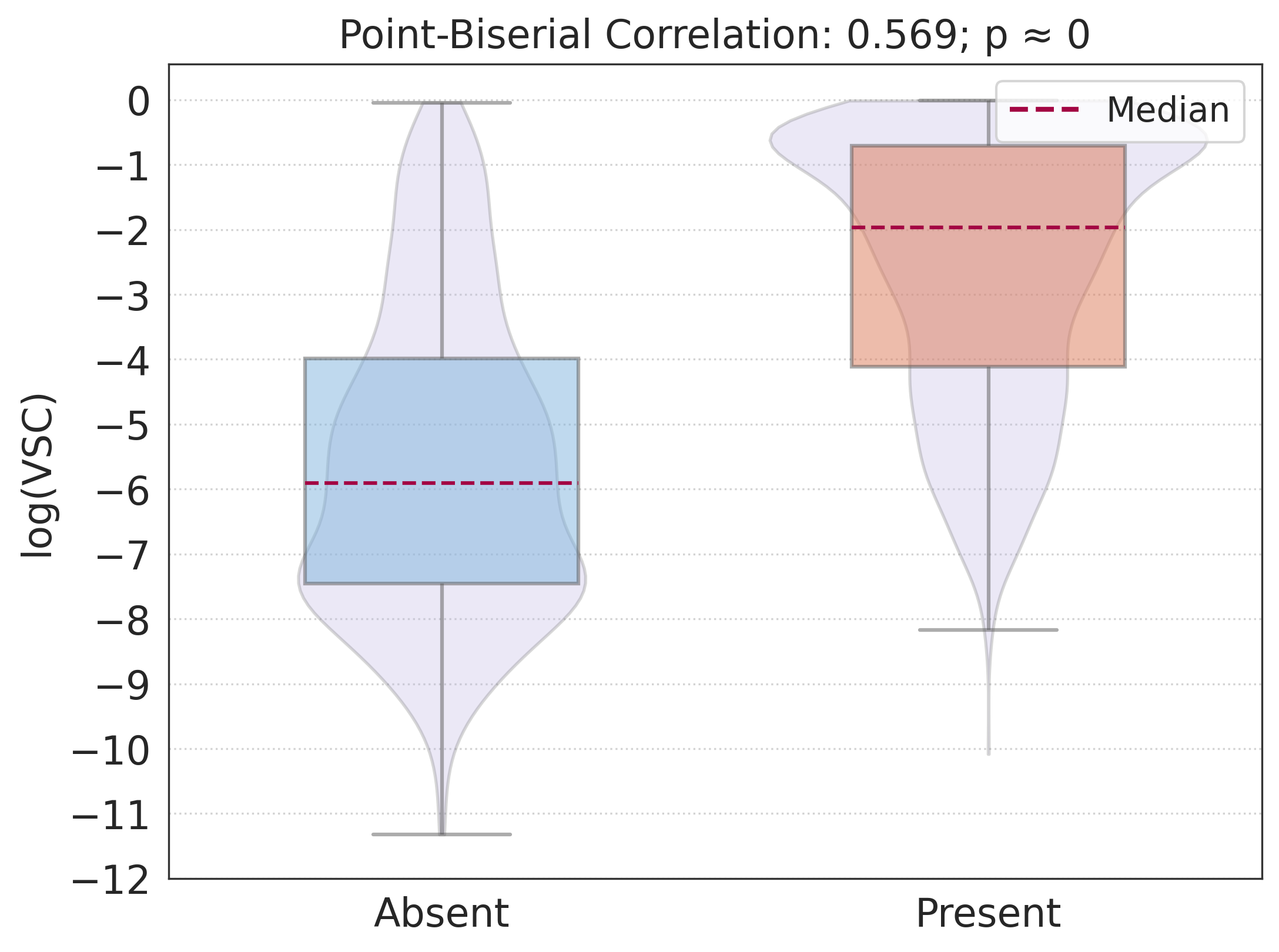}
    \caption{Analysis of visual semantic confidence.}
    \label{fig:vsc_cor}
\end{figure}

\begin{figure}
  \centering
  \begin{subfigure}{0.49\linewidth}
    \includegraphics[width=\linewidth]{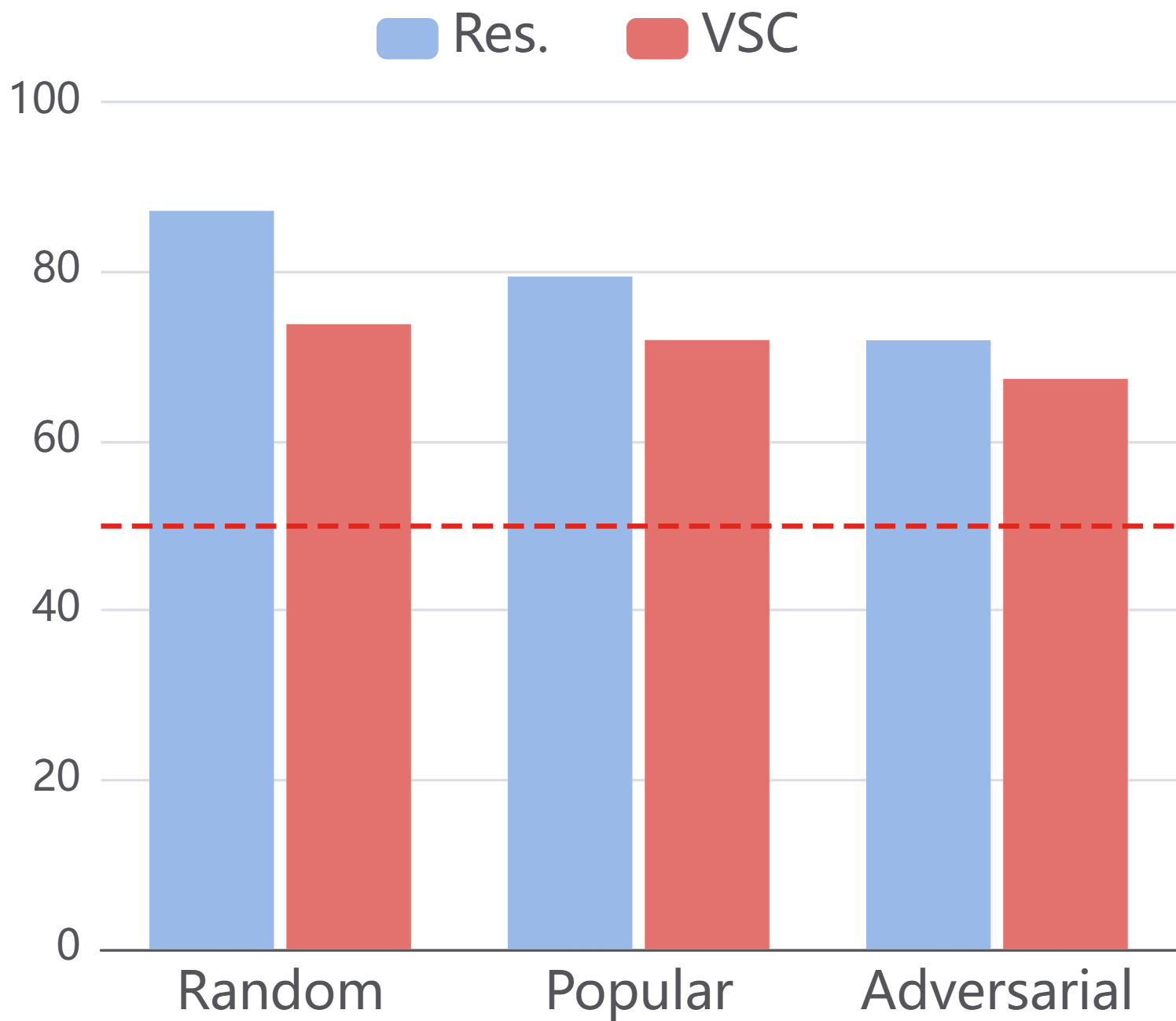}
  \end{subfigure}
  \hfill
  \begin{subfigure}{0.49\linewidth}
    \includegraphics[width=\linewidth]{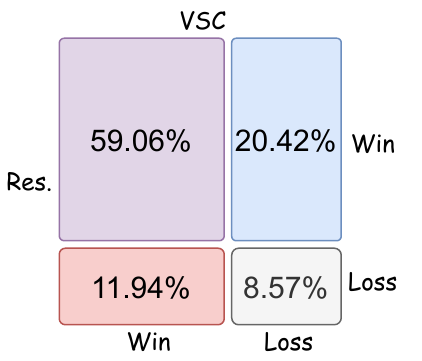}
  \end{subfigure}
  \caption{Accuracy comparison of LLaVA-1.5-7B's response (\emph{Res.}) and its \emph{VSC} on the POPE benchmark.
   The threshold of VSC for determining the presence of an object is set to $\log[c(O)] > -2.5$, a value obtained via grid search optimizing the F1 score on the Random subset of the COCO dataset.
  }
  \label{fig:vsc_pfm}
\end{figure}

The VSC of a visual token $v_i$ with respect to an object $O$ is defined as $c_{v_i}(O) = \mathrm{softmax}[\mathrm{logit}_{v_i}(O)]$.
Given the autoregressive nature of MLLMs, we approximate this confidence using the first token $o_0$ of the tokenized object $O$ (\emph{i.e.}, $c_{v_i}(O) \approx c_{v_i}(o_0)$).

We employ an existence-type VQA task to analyze the relationship between the VSC and the object. We represent the semantic confidence of the entire input image with respect to object $O$ by applying max pooling over all visual tokens:
\begin{equation}
    c(O) = \max c_{v_i}(o_0) .
\end{equation}
The analysis results are shown in \Cref{fig:vsc_cor}. As observed, VSC exhibits a strong correlation with object presence: objects appearing in the image consistently produce higher semantic confidence scores. This finding indicates that VSC effectively captures the visual semantics encoded in the visual tokens.

VSC serves as an intuitive indicator of an MLLM's capacity to interpret visual information. We further examine the consistency between VSC and the model's responses, with results presented in \Cref{fig:vsc_pfm}. Although object-level judgments derived from VSC are less accurate than the model's own responses, they still demonstrate a correct preference, significantly exceeding 50\% accuracy. In addition, we observe that the understanding reflected by VSC does not always fully align with the model's responses, revealing a certain degree of preference discrepancy. These findings suggest that an MLLM's visual understanding is not yet fully exploited, and that the VSC mechanism may be used to enhance the model's visual perception.

\begin{figure}
    \centering
    \includegraphics[width=0.9\linewidth]{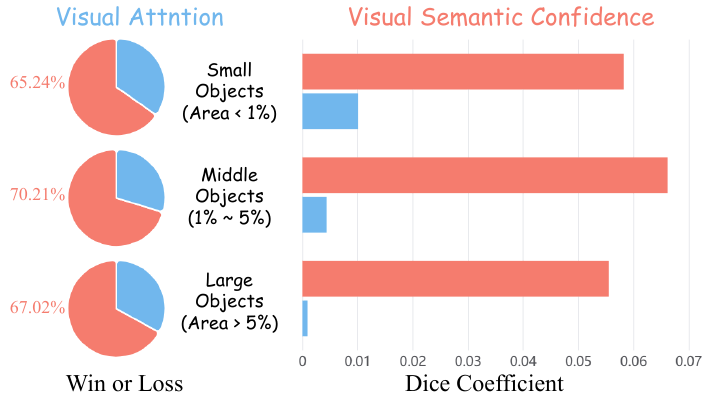}
    \caption{Comparison of visual grounding performance between VSC and the visual attention mechanism of LLaVA-1.5-7B. The visual attention maps are extracted from layer 14, following the prior work~\cite{ViCrop}.}
    \label{fig:vsc-vattn}
\end{figure}

Visual tokens associated with an object exhibit higher VSC values with respect to that object. Leveraging this property, we obtain the visual grounding of the object as:
\begin{equation}\label{eq:G}
    \bm{G}_O=\mathrm{Norm}[\{c_{v_i}(o_0)\}_{i=1}^m] \in \mathbb{R}^{m} ,
\end{equation}
where $m$ denotes the number of visual tokens, and $\mathrm{Norm}(\cdot)$ represents sum normalization.
To quantify the grounding performance of VSC, we employ the Dice Coefficient \cite{SECOND}:
\begin{equation}
    D = \frac{2\sum^m_{i} c_i g_i}{\sum^m_{i} c_i + \sum^m_{i} g_i},
\end{equation}
where $g_i \in [0,1]$ denotes the overlap coefficient between visual token $v_i$ and the object's mask annotation.
We randomly select 500 images from MSCOCO to evaluate the grounding capability of VSC and compare it with visual attention.
The results are presented in \Cref{fig:vsc-vattn}. We observe that VSC demonstrates superior grounding capability compared to visual attention.
Visual attention suffers from the ``sink'' phenomenon \cite{VAR}, which limits its grounding ability—particularly for large objects, where it tends to focus only on local regions.
In contrast, VSC demonstrates substantially stronger visual grounding performance, exhibiting consistent behavior across objects of different scales.
This token-level grounding capability of VSC also provides an indicator of the most informative visual tokens.

\subsection{Object-Agnostic Grounding}\label{sec:object-agnostic}

In VQA tasks, a specific object is often explicitly referenced, enabling the use of VSC for precise object grounding. However, this strategy is not applicable to image captioning, which is inherently object‑agnostic.
Moreover, in image captioning, the model should not concentrate on a single object; instead, it must attend to all visual information in the image—particularly to visual tokens that encode richer visual content.

\begin{figure}
    \centering
    \includegraphics[width=\linewidth]{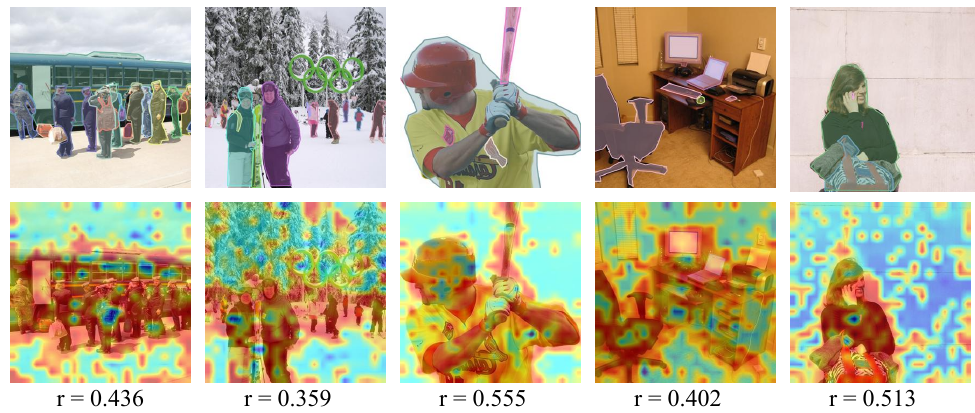}
    \caption{Point-biserial correlation between visual semantic salience and object masks. }
    \label{fig:vss}
\end{figure}
We assume that visual tokens rich in visual information should exhibit more definite visual semantics.
We employ output uncertainty \cite{LookTwice,CLVS} to measure the semantic salience in visual tokens:
\begin{equation}
    c_{v_i} = -\sum_k c_{v_i}(w_k)\cdot \log c_{v_i}(w_k) / \log K ,
\end{equation}
where $\{w_k\}$ are Top-$K$ tokens in $\mathrm{logit}_{v_i}$.
To distinguish this approach from VSC, we refer to it as Visual Semantic Salience (VSS).

The cases in \Cref{fig:vss} demonstrate the effectiveness of VSS. We observe that tokens corresponding to distinct objects exhibit higher VSS values, whereas semantically insignificant background regions yield lower VSS. Leveraging this property, we construct visual grounding for all objects present in the image.

\section{Method}
\label{sec:method}

We attribute hallucinations in MLLMs to two primary factors:
(1) \emph{Underutilization of extracted visual features}: The model may fail to generate a correct response even when it successfully captures object features with high VSC.
(2) \emph{Suboptimal visual attention}: Although visual attention is the main pathway through which the model interprets an input image, it often struggles to accurately focus on the most relevant visual regions.
Motivated by these observations,
we propose Vision-Guided Attention (VGA), which leverages VSC-derived grounding to direct visual attention toward the most informative tokens.

\subsection{Vision-Guided Attention}

Due to the autoregressive nature of MLLMs, we can pre‑process the visual tokens to compute the VSC and then leverage the caching mechanism to continue generating the response, ensuring that each token undergoes only a single forward pass. 
We first use NLP tools to extract the objects mentioned in the question and then construct visual grounding through the VSC mechanism. When multiple objects appear in the question, we apply max pooling over the grounding corresponding to all extracted objects:
\begin{equation}
    \bm{G} = \max_j \bm{G}_{O_j} \in \mathbb{R}^m ,
\end{equation}
where $\bm{G}_{O_j}$ is defined in \Cref{eq:G}, and $m$ is the number of visual tokens, satisfying $m=e-s$.
We use $\bm{G}$ to guide the model's visual attention:
\begin{equation}
    \hat{\bm{\alpha}}_{h,-1,s:e} = \bm{\alpha}_{h,-1,s:e} + \beta \cdot \bm{G} ,
\end{equation}
where $\beta$ is a hyperparameter, and $\bm{\alpha} \in \mathbb{R}^{H\times L\times L}$ denotes the attention weights, with $H$ representing the number of attention heads and $L$ the sequence length. 
The model then applies the guided attention weights to compute self‑attention:
\begin{equation}
    \hat{\bm{z}} = \hat{\bm{\alpha}}\bm{V},
\end{equation}
where $\bm{V}$ denotes the value embeddings, and $\hat{\bm{z}}$ represents the self‑attention output embeddings.

Attention acceleration techniques such as FlashAttention provide high computational efficiency, making them the preferred choice in most MLLM applications. However, these methods do not explicitly output attention weights (\emph{i.e.} $\bm{\alpha}$), rendering various attention-based optimization approaches incompatible. In contrast, VGA optimizes attention using visual grounding derived from VSC; since it does not require the computation of attention weights, it remains fully compatible with FlashAttention.

VGA can be optimized using the associative property of addition:
\begin{equation}
   \hat{\bm{z}} = (\bm{\alpha}+ \beta \cdot \bm{G})V = \bm{z} +  \beta \cdot \Delta \bm{z} ,
\end{equation}
where $\bm{z}$ denotes the output embeddings of vanilla self-attention

\subsection{Attention Heads Balancing}
The multi-head attention mechanism inherently promotes spontaneous specialization among attention heads, resulting in certain heads naturally developing visual functionalities. To prevent disruption to the model's intrinsic visual capabilities, we balance the guidance strength across attention heads: weaker guidance is applied to visually specialized heads, while stronger guidance is employed for non-visual heads.
The update for the attention mechanism is formulated as follows:
\begin{equation}\label{eq:update}
   \hat{\bm{z}} = \bm{z} +  \beta \cdot  \bm{\gamma} \cdot \Delta \bm{z} ,
\end{equation}
where $\bm{\gamma} \in \mathbb{R}^H $ represents the balancing coefficients across the attention heads.
To approximate the discrepancy in visual functionality across all attention heads, we utilize the similarity between $\bm{z}$ and $\Delta \bm{z}$:
\begin{equation}
    \bm{\gamma}' = \mathrm{Norm}(\mathrm{sim}(\bm{z}, \Delta \bm{z})) \in \mathbb{R}^H ,
\end{equation}
where $\mathrm{Norm}(\cdot)$ is the sum normalization and $\mathrm{sim}(\cdot,\cdot) \in [0,1]$ is the cosine similarity.
A higher similarity value for a particular head signifies a more pronounced visual functionality.
Finally, we scale this discrepancy to have a mean of $1$ and then invert it to obtain the final balancing coefficient 
\begin{equation} 
    \bm{\gamma} = \mathrm{ReLU}(2-H \cdot \bm{\gamma}') ,
\end{equation}
where $H$ is the number of attention heads.

\subsection{Programmed Vision-Guidance}

In VQA tasks, visual demands are typically static and remain unchanged throughout the inference process. In contrast, image captioning involves dynamically evolving visual demands as the description is being generated. Consequently, a constant vision grounding mechanism is insufficient to adequately meet the varying visual requirements of image captioning tasks.

To enable visual grounding that dynamically adapts to these changing visual demands during the generation process, we propose the Programmed Visual Grounding (PVG) mechanism. Specifically, during text generation, we dynamically adjust the visual grounding by suppressing regions associated with content that has already been generated.

The adjustment of the visual grounding is performed as follows:
\begin{equation}
    \bm{G}_{t+1} = (1+\lambda)  \bm{G}_t - \lambda \bm{G}_w ,
\end{equation}
where $w$ is the token generated at the current step and $\lambda$ is a hyperparameter controlling the change intensity.
This adjustment mechanism effectively shifts the vision grounding of guidance away from regions associated with already generated content $\bm{G}_w$ and toward regions relevant to the yet-to-be-generated content.
Subsequently, the visual grounding is sum-normalized as follows:
\begin{equation}
    \bm{G}'_{t+1} = \mathrm{Norm}(\mathrm{ReLU}(\bm{G}_{t+1})) .
\end{equation}

In image captioning tasks, as more content is generated, the guidance shifts toward less informative regions via the PVG mechanism, and the model's dependency on visual information gradually decreases \cite{TARAC}. Accordingly, the vision-guidance should be progressively weakened.
The attention update mechanism presented in \Cref{eq:update} is therefore modified to incorporate this decay:
\begin{equation}
    \hat{\bm{z}} = \bm{z} + ||\bm{G}||_0 \cdot \bm{\gamma} \cdot \beta \cdot \Delta \bm{z} ,
\end{equation}
where $||\bm{G}||_0$ is the L0 norm of $\bm{G}$.

\begin{table*}
    \centering
    \caption{
    Results on POPE. The results are reported as the average performance across the MSCOCO, A-OKVQA, and GQA datasets.
    \textbf{Bolded} values indicates the best results and \underline{underlined} values indicates the second-best results.
    \emph{Acc.} stands for Accuracy.
    }\label{tab:pope}
    \begin{tabular}{l|l|cc|cc|cc|cc}
       \toprule
& \multirow{2}{*}{Method}
       &  \multicolumn{2}{c|}{LLaVA-7B} &  \multicolumn{2}{c|}{LLaVA-13B} &  \multicolumn{2}{c|}{LLaVA-Next} &  \multicolumn{2}{c}{Qwen2.5-VL}\\
        \cmidrule(lr){3-4}\cmidrule(lr){5-6}\cmidrule(lr){7-8}\cmidrule(lr){9-10}
       &  &  Acc. $\uparrow$  &  F1$\uparrow$ & Acc. $\uparrow$ &  F1 $\uparrow$&  Acc. $\uparrow$&  F1 $\uparrow$ &  Acc. $\uparrow$ &  F1 $\uparrow$\\
       \midrule
       \multirow{6}{*}{Random} 
       & Vanilla &  87.12 & 87.94 & 84.31 & 85.96 & 87.36 & 85.90 & 89.88 & 89.10    \\
       & PAI &    86.20 & 87.20 & 85.36 & 86.71 & 87.71 & 86.34 & \underline{90.85} & \underline{90.31}     \\
       & PAI$_\mathrm{CD}$ &  \underline{87.27} & \underline{87.98} & 86.10 & 87.24 & - & - & - & -   \\
       & VAF &    84.88 & 86.34 & 82.42 & 84.62 & \underline{89.31} & \underline{88.45} & 90.57 & 89.87  \\
       & TARAC &  86.99 & \underline{87.80} & \underline{87.02} & 87.80 & 86.13 & 84.26 & 89.79 & 89.01       \\
       \rowcolor{gray!20}
         \cellcolor{white}
       & VGA &   \textbf{89.28} & \textbf{89.41} & \textbf{88.58} & \textbf{89.07} & \textbf{89.91} & \textbf{89.21} & \textbf{91.08} & \textbf{90.57}   \\
        \midrule
       \multirow{6}{*}{Popular} 
       & Vanilla &    79.38 & 82.14 & 78.64 & 81.87 & 85.45 & 84.14 & 87.88 & 87.24   \\
       & PAI &   78.45 & 81.56 & 79.72 & 82.56 & 85.95 & 84.70 & \underline{87.93} & \underline{87.65}         \\
       & PAI$_\mathrm{CD}$ &   \underline{80.03} & \underline{82.51} & \underline{80.86} & \underline{83.31} & - & - & - & -  \\
       & VAF &   77.13 & 80.82 & 77.10 & 80.91 & \underline{86.58} & \underline{85.95} & 87.90 & 87.41   \\
       & TARAC &    79.37 & 82.10 & 79.40 & 82.10 & 83.94 & 82.25 & 87.88 & 87.22       \\
       \rowcolor{gray!20}
         \cellcolor{white}
       & VGA &   \textbf{83.51} & \textbf{84.73} & \textbf{84.15} & \textbf{85.50} & \textbf{86.62} & \textbf{86.24} & \textbf{88.06} & \textbf{87.81}       \\
       \midrule
       \multirow{6}{*}{Adversarial} 
       & Vanilla &   71.85 & 77.04 & 71.58 & 77.23 & 82.31 & 81.40 & \underline{84.82} & 84.57    \\
       & PAI &   70.83 & 76.46 & 72.38 & 77.64 & 82.64 & 81.82 & 84.60 & \underline{84.82}        \\
       & PAI$_\mathrm{CD}$ &  \underline{72.61} & \underline{77.38} & \underline{73.44} & \underline{78.26} & - & - & - & -    \\
       & VAF &   69.59 & 75.94 & 70.08 & 76.41 & \textbf{83.14} & \underline{83.00} & 84.80 & 84.74     \\
       & TARAC &   71.92 & 77.03 & 71.95 & 77.03 & 81.04 & 79.73 & 84.72 & 84.47      \\
       \rowcolor{gray!20}
         \cellcolor{white}
       & VGA &   \textbf{77.02} & \textbf{79.88} & \textbf{77.01} & \textbf{80.28} & \underline{83.06} & \textbf{83.22} & \textbf{84.86} & \textbf{85.09}      \\
       
       \bottomrule
    \end{tabular}
\end{table*}

\begin{table*}
    \centering
    \caption{
    Results on CHAIR. Ci and Cs represent the CHAIRi and CHAIRs metric, respectively.
    }\label{tab:chair}
    \begin{tabular}{l|ccc|ccc|ccc|ccc}
       \toprule
        \multirow{2}{*}{Method} &
         \multicolumn{3}{c|}{LLaVA-7B} &  \multicolumn{3}{c|}{LLaVA-13B} &  \multicolumn{3}{c|}{LLaVA-Next} &  \multicolumn{3}{c}{Qwen2.5-VL}\\
       \cmidrule(lr){2-4}\cmidrule(lr){5-7}\cmidrule(lr){8-10}\cmidrule(lr){11-13}
        & Cs $\downarrow$ & Ci $\downarrow$ &  F1 $\uparrow$ & Cs $\downarrow$ & Ci  $\downarrow$ &  F1 $\uparrow$  & Cs $\downarrow$ & Ci $\downarrow$ &  F1 $\uparrow$ & Cs $\downarrow$ & Ci $\downarrow$ &  F1 $\uparrow$\\
        \midrule
        Vanilla &    
        52.2 & 13.9 & 76.1    & 
        52.6 & 14.1 & 76.4 &   
        34.2 & 9.3 & 72.2  &    
        39.2 & \underline{9.7} & \underline{75.0} \\
        
        PAI &  
        33.6 & 9.0 & \textbf{77.1} &    
        40.2 & 10.8 & 76.9  &    
        34.0 & 8.9 & \textbf{73.7}  &     
        \underline{36.0} & 10.3 & 71.9   \\
        
        PAI$_\mathrm{CD}$ &  
        \underline{30.8} & \textbf{8.1} & \underline{76.3}   &  
        \underline{37.0} & \underline{9.9} & \textbf{77.3}  &   
        -  &  -   & - &  
        -   & - & - \\
        
        VAF &   
        52.4 & 14.6 & 75.3    &  
        49.6 & 13.6 & 76.5  &  
        32.8 & \underline{8.8} & \underline{72.6}  &     
        38.2 & 9.8 & 74.8 \\
        
        TARAC &   
        43.0 & 11.2 & 75.1   &   
        42.6 & 11.6 & 76.9 &  
        \underline{29.8} & \textbf{8.0} & 71.5  &   
        39.8 & 9.8 & 74.6 \\
        
        \rowcolor{gray!20}
        VGA &   
        \textbf{29.8} & \underline{8.2} & \underline{76.3}   &  
        \textbf{30.0} & \textbf{8.2} & \underline{77.1}   &   
        \textbf{28.6} & \textbf{8.0} & 71.6  &     
        \textbf{35.4} & \textbf{9.4} & \textbf{75.1}   \\
        
       \bottomrule
    \end{tabular}
\end{table*}

\section{Experiments}
\label{sec:exp}

\subsection{Experimental Setup}

\paragraph{Benchmarks \& Baselines.}
Our method is evaluated on three hallucination benchmarks: POPE~\cite{POPE} (VQA), CHAIR~\cite{chair} (image captioning), and AMBER~\cite{AMBER} (hybrid).
We select three recent visual attention-based dehallucination methods for comparative analysis: PAI~\cite{PAI}, VAF~\cite{VAF}, TARAC~\cite{TARAC}.
We denote the variants of PAI with and without contrastive decoding as PAI$_\mathrm{CD}$ and PAI, respectively.  

\paragraph{Implementation Details.}
Following the established practice of PAI, we determine the starting layer for VGA based on the \emph{BOS} attention. Specifically, the starting layers are set as follows (with layers numbered from 0): layer 2 for LLaVA-1.5, layer 0 for LLaVA-Next, and layer 4 for Qwen2.5-VL. We apply VGA up to an intermediate layer of the model: specifically, layer 24 for LLaVA-1.5-13B and layer 16 for all other models.
We adopt the default settings of $\beta=0.2$ and $\lambda=0.02$.
For the larger model (LLaVA-1.5-13B) and visually simpler tasks (POPE), we increase $\gamma$ to $0.25$ to apply stronger vision guidance.
Stanza~\cite{stanza} is utilized to extract objects mentioned in the question.
We employ greedy decoding for next token prediction and set the maximum generation length to 512.

Detailed experimental setups are provided in the Appendix~\ref{app:exp setup}.

\subsection{Main Results}

\begin{table*}
    \centering
    \caption{
    Results on AMBER. The AMBER metric is calculated as $ (1-\text{CHAIR}+\text{F1})/2 $.
    }\label{tab:amber}
    \begin{tabular}{ll|cccc|cccc|c}
       \toprule
      MLLM & Method & CHAIR $\downarrow$ & Cover $\uparrow$ &  Hal $\downarrow$& Cog $\downarrow$ & Acc. $\uparrow$&  Prec. $\uparrow$ & Rec. $\uparrow$ & F1 $\uparrow$ &  AMBER $\uparrow$ \\
       \midrule
     \multirow{6}{*}{LLaVA-7B}
     &   Vanilla &  7.1 & \textbf{50.7} & 32.5 & 3.8 & 72.1 & 92.2 & 63.0 & 74.8 & 34.35 \\
     &  PAI &   5.8 & 48.8 & 27.5 & 2.5 & 70.3 & \textbf{94.4} & 58.8 & 72.5 & 33.85  \\
     &  PAI$_\mathrm{CD}$ &  \underline{5.1} & 48.4 & \underline{25.7} & \underline{2.1} & \underline{73.7} & \underline{93.1} & \underline{65.2} & \underline{76.7} & \underline{36.30} \\
      &  VAF &   6.9 & \textbf{50.7} & 33.0 & 3.5 & 68.7 & 92.6 & 57.4 & 70.9 & 32.50    \\
     &  TARAC &   5.8 & \underline{49.3} & 28.5 & 2.9 & 72.0 & 92.8 & 62.6 & 74.8 & 35.00  \\
      \rowcolor{gray!20}
        \cellcolor{white}
      &  VGA &   \textbf{4.5} & 48.8 & \textbf{21.6} & \textbf{1.9} & \textbf{76.2} & \underline{93.1} & \textbf{69.3} & \textbf{79.5} & \textbf{38.00}  \\
      \midrule
     \multirow{5}{*}{LLaVA-Next}
     &   Vanilla &    7.8 & 63.1 & 46.1 & 3.9 & 85.9 & 86.5 & 93.4 & 89.8 & 41.50 \\
     &  PAI &    \underline{7.1} & \textbf{64.2} & 42.2 & \underline{3.2} & \underline{86.4} & 87.4 & 93.0 & \underline{90.1} & \underline{42.00}   \\
      &  VAF &     7.8 & \underline{63.2} & 45.4 & 3.9 & 86.0 & \underline{87.7} & 91.9 & 89.7 & 41.45    \\
     &  TARAC &   \textbf{6.7} & 60.3 & \textbf{37.5} & 3.4 & 85.8 & 86.3 & \textbf{93.5} & 89.8 & \textbf{42.05}   \\
      \rowcolor{gray!20}
        \cellcolor{white}
      &  VGA &   \underline{7.1} & 62.2 & \underline{41.4} & \textbf{3.1} & \textbf{86.6} & \textbf{88.4} & 92.0 & \textbf{90.2} & \textbf{42.05}    \\
      \midrule
     \multirow{5}{*}{Qwen2.5-VL}
     &   Vanilla &   \underline{4.8} & \underline{64.8} & 25.9 & 1.5 & 86.5 & 87.3 & \textbf{93.2} & 90.1 & 43.15 \\
     &  PAI &    7.1 & 60.3 & 34.7 & 1.6 & 86.0 & \underline{88.1} & 91.2 & 89.6 & 41.75  \\
      &  VAF &     \textbf{4.7} & 64.2 & \underline{25.1} & \underline{1.4} & 85.7 & 87.1 & 92.1 & 89.5 & 42.90   \\
     &  TARAC &   \underline{4.8} & \textbf{65.2} & 26.5 & 1.5 & \underline{86.6} & 87.8 & \underline{92.7} & \underline{90.2} & \underline{43.20}   \\
      \rowcolor{gray!20}
        \cellcolor{white}
      &  VGA &  \textbf{4.7} & 61.8 & \textbf{24.2} & \textbf{1.3} & \textbf{87.1} & \textbf{88.5} & 92.5 & \textbf{90.5} & \textbf{43.40}   \\
       \bottomrule
    \end{tabular}
\end{table*}

POPE results are presented in \Cref{tab:pope}. Our proposed method, VGA, achieves significant improvements, clearly demonstrating its advantage in mitigating object hallucinations. The results on CHAIR, shown in \Cref{tab:chair}, demonstrate that VGA also performs well on the image captioning task, achieving the lowest overall hallucination rate across all evaluated MLLMs while maintaining a comparable F1 score. \Cref{tab:amber} presents the results on AMBER, where VGA achieves the best overall performance, further validating the effectiveness of VGA.

We provide a more comprehensive discussion of the experimental results in the Appendix~\ref{app:main results}.

\subsection{Programmed Vision-Guidance for Image Captioning}
\begin{figure}
    \centering
    \includegraphics[width=\linewidth]{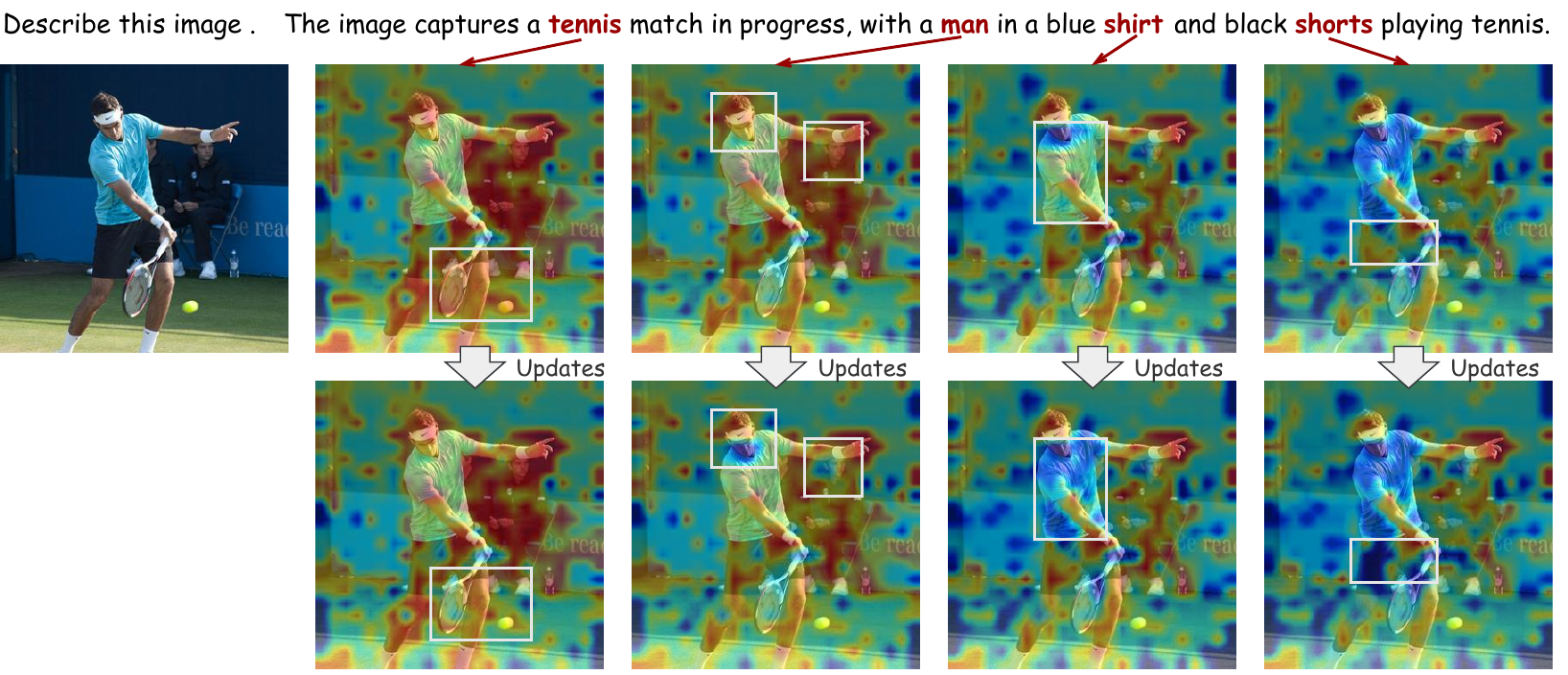}
    \caption{Programmed Vision-Guidance. We set $\gamma$ to 0.1 in this case to better illustrate the dynamic process. }
    \label{fig:guidance}
\end{figure}
For the image captioning task, we first construct visual grounding based on VSS. Subsequently, we dynamically adjust this guidance during generation by suppressing already-described visual regions—a process we term programmed vision-guidance.
As illustrated in \Cref{fig:guidance}, the precise grounding capability enabled by VSC facilitates the accurate editing of vision-guidance. This mechanism effectively steers the model to focus more on previously undescribed regions, promoting comprehensive and non-redundant caption generation.

\begin{table}
    \centering
        \caption{\emph{PVG} represents Programmed Vision-Guidance. \emph{Reversed VSS} denotes initial vision guidance by $1-\bm{G}$ in \cref{eq:G}.
        }\label{tab:pvg}
    \begin{tabular}{l|cccc}
    \toprule
       Setting &  Cs & Ci  &  R &  F1\\
         \midrule
        VGA &  29.8 & 8.2 & \textbf{70.4} & \textbf{76.3}    \\
     Reversed VSS & 30.6 & 8.6 & 69.5 & 75.9  \\
       \emph{w/o} PVG & \textbf{22.2} & \textbf{5.6} & 65.0   & 74.5  \\
      \bottomrule
    \end{tabular}
\end{table}

We conduct an ablation study comparing PVG with static VSS-based grounding, with the results presented in \Cref{tab:pvg}.
First, guiding the model to focus on semantically rich regions effectively improves image captioning quality by encouraging the generation of more meaningful and informative content. Conversely, increasing attention to low-semantic regions not only leads to a rise in hallucinations but also results in a decline in recall, as the model generates less relevant or redundant descriptions.
Moreover, models operating without dynamic adjustment of visual guidance tend to produce more concise descriptions. While this static approach may reduce hallucination rates, it concurrently leads to the omission of important visual details.
In summary, PVG significantly enhances MLLM performance in image captioning. It achieves this by balancing detail preservation and hallucination suppression through adaptive visual grounding, thereby mitigating the trade-off inherent in static grounding methods.

\begin{figure}
    \centering
    \includegraphics[width=0.6\linewidth]{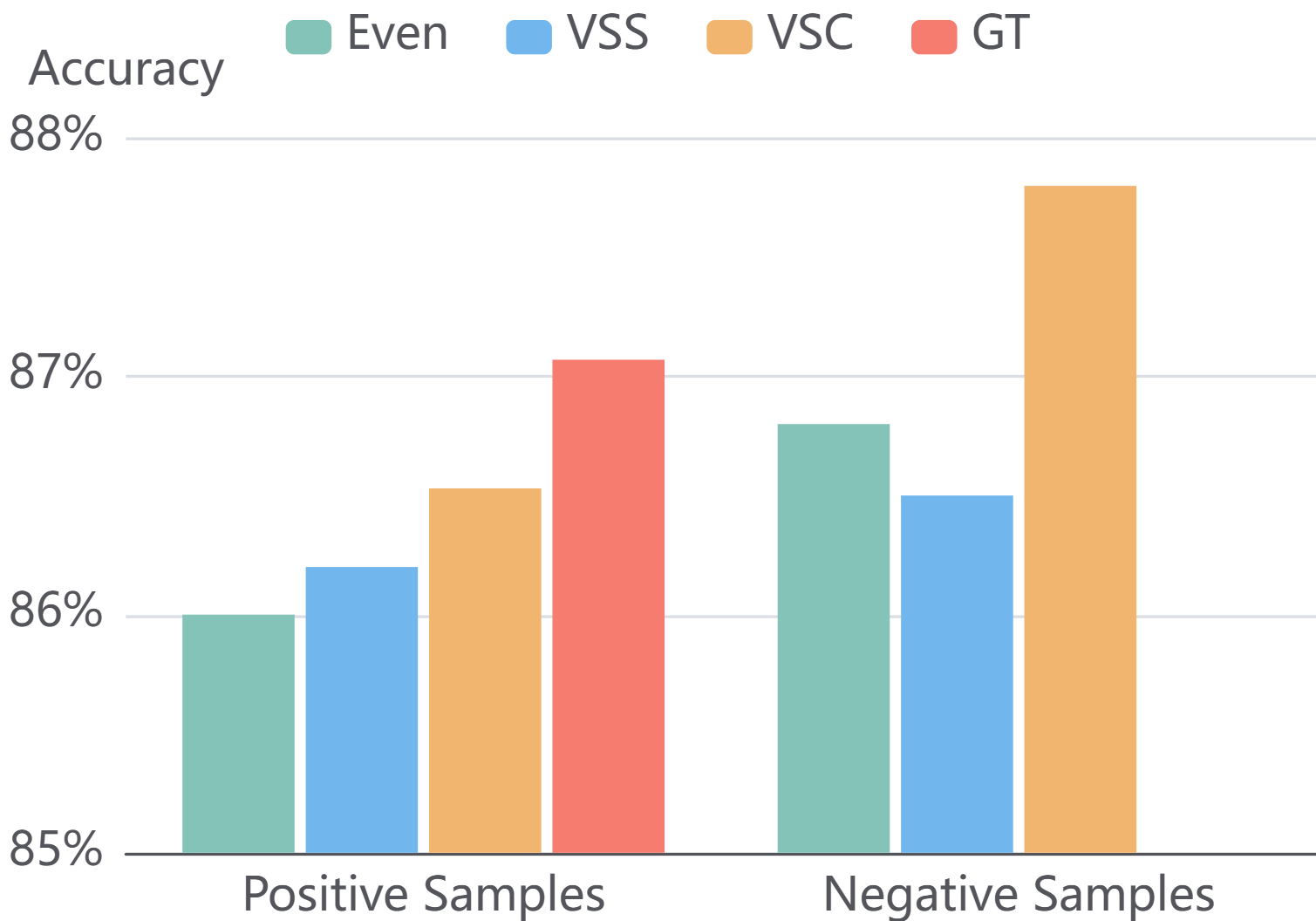}
    \caption{Accuracy on positive and negative samples in POPE with different vision-guidance.
    \emph{Even} denotes uniform attention allocation across all visual tokens, while \emph{GT} refers to the use of ground-truth grounding.}
    \label{fig:pos_neg}
\end{figure}
\begin{figure}
    \centering
    \includegraphics[width=\linewidth]{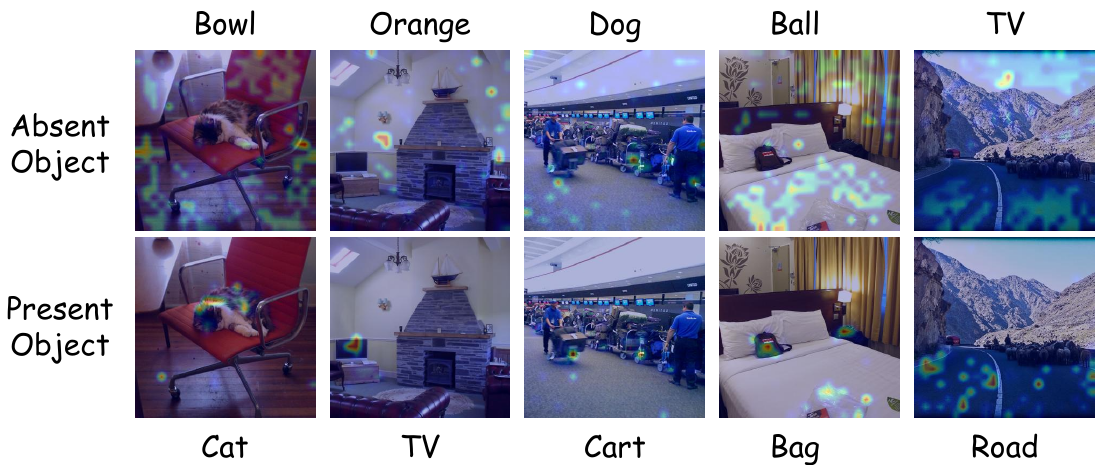}
    \caption{VSC grounding for absent and present objects.}
    \label{fig:abs_pres}
\end{figure}

\subsection{Vision-Guidance Enhancing MLLM's Visual Perception}

VGA employs VSC grounding to guide visual attention, leading to significant performance gains on the POPE benchmark.
To better understand the role of grounding, we further investigate two critical questions:
(1) \emph{Why does grounding for absent objects still effectively improve the model's object perception?}
(2) \emph{What will happen when VSC grounding is unreliable?}

\subsubsection{Analysis of Negative Samples}
\Cref{fig:pos_neg} illustrates the effects of different visual grounding strategies on both positive and negative samples.
First, in positive samples—where the queried object is present—the model's object perception consistently improves as the visual grounding becomes more precise.
This demonstrates that providing precise visual grounding effectively enhances MLLM's overall visual understanding.
Moreover, in negative samples—queries about objects absent from the image—the application of VSC still significantly improves the model's ability to correctly reject the presence of such objects.

\Cref{fig:abs_pres} presents the VSC-based grounding maps for both present and absent objects.
We find that VSC enhances the model's perception for absent objects by redirecting attention toward low-semantic regions (\emph{i.e.}, background or uninformative areas), allowing the model to easily infer the object's absence. In contrast, when guided by VSS-based grounding—which directs attention toward visually richer regions—the model is more prone to misinterpretation, leading to hallucinatory responses.

To quantitatively analyze VSC grounding of negative samples, we define high-semantic regions (above-mean) based on the VSS distribution. We introduce the Informative Coefficient (IC), defined as the sum of VSC weights (sum-normalized) over these high-semantic regions.
The average IC is 0.64 for positive samples and 0.42 for negative samples, indicating that visual guidance in negative samples concentrates on low-semantic regions. Moreover, false positives (FPs) exhibit an average IC of 0.57, whereas true negatives (TNs) show an average of 0.41. After randomly sampling TNs to match the FP count, we analyze their correlation; the average Pearson correlation between IC and FP is $r = 0.414\pm0.019$. These results demonstrate that VSC effectively shifts attention away from high-semantic regions in negative contexts, thereby reducing FPs.

\subsubsection{Spurious VSC Grounding}
We analyze the performance of VGA under conditions of imprecise visual grounding. On positive samples, VGA employing spurious VSC grounding with a low Dice Coefficient (DICE $<$ 0.01) achieves an accuracy of 88.5\%, compared to 87.1\% for VGA with uniform guidance.
We attribute the beneficial effect of spurious grounding to information flow: final visual hidden states inherently incorporate features from preceding visual tokens, resulting in a misalignment between vision grounding and feature grounding. Therefore, although VSC localization may be visually suboptimal in some cases, it still provides correct positional preferences.

We further analyze the failure cases of VGA. We observe that true positives exhibit significantly higher average DICE scores (0.0475) compared to false negatives (0.0006).
In these failure cases, the visual grounding constructed by VSC exhibits extremely low precision.
This is attributed to the model's inability to effectively extract visual features, which renders the VSC derived from degraded visual logits ineffective. In the absence of external cues, failures in visual feature extraction inevitably lead to erroneous understanding.
Nevertheless, the consistent performance improvements across various baselines indicate that the VSC mechanism possesses robust visual localization capabilities.

In short, precise vision guidance is crucial for the model's visual understanding, and the VSC mechanism provides reliable guidance.

\subsection{The Evaluation of Latency}
Both the construction of VSC grounding and the update of vision guidance via PVG utilize pre-computed visual logits (fixed during inference), requiring no additional forward passes.
We evaluate Time to First Token (TTFT) using LLaVA-1.5-7B on the AMBER benchmark, where VGA incurs only a 4.36\% increase. Additionally, we evaluate Time Per Output Token (TPOT) on CHAIR, where VGA introduces a 6.01\% overhead.
In short, our method is well-compatible with the model's autoregressive nature, incurring only minimal latency overhead.

\section{Conclusion}

In this work, we propose a Visual Semantic Confidence mechanism to construct accurate visual grounding without relying on external tools. Building upon this, we introduce Vision-Guided Attention, a novel mechanism that leverages the derived visual grounding to steer the model's visual attention toward the most informative regions, thereby effectively mitigating hallucinations in MLLMs. Moreover, it requires only a single forward pass per token, incurs negligible additional inference cost, and is compatible with FlashAttention.

\section*{Acknowledgments}
We thank all reviewers for their detailed reviews and valuable comments.
This work is supported by Zhongguancun Academy Project No.20240103.

{
    \small
    \bibliographystyle{ieeenat_fullname}
    \bibliography{main}
}

\clearpage
\setcounter{page}{1}
\maketitlesupplementary

\renewcommand\thesection{\Alph{section}}
\setcounter{section}{0}
\setcounter{figure}{9}
\setcounter{table}{4}
\setcounter{equation}{15}

\section{Detailed Experimental Setup}\label{app:exp setup}
\paragraph{Evaluated MLLMs.}
We evaluate the effectiveness of our proposed method, Vision-Guided Attention (VGA), on three representative MLLMs: \textbf{LLaVA-1.5}~\cite{llava15}, \textbf{LLaVA-Next}~\cite{llava-next}, and \textbf{Qwen2.5-VL}~\cite{qwen25-vl}.
LLaVA-1.5 is available in two sizes: 7B and 13B. Both LLaVA-Next and Qwen2.5-VL are 8B-scale

\subsection{Benchmarks}
In order to validate the effectiveness of our method in mitigating hallucination, we evaluate its performance on three widely-used multimodal hallucination benchmarks: one generative benchmark (CHAIR \cite{chair}), one discriminative benchmark (POPE \cite{POPE}), and one hybrid benchmark (AMBER \cite{AMBER}).

\paragraph{CHAIR.} CHAIR evaluates the proportion of hallucinated objects, which are generated by the model but not present in the reference annotations. Following prior works, we randomly select 500 images from the MSCOCO \cite{coco} dataset as the test set. We additionally select another 500 images to construct a validation set for hyperparameter tuning. 
This benchmark includes two metrics: CHAIRs and CHAIRi, defined as follows:
\begin{equation}
\begin{aligned}
   \text{CHAIRs} &= \frac{|\text{Hallucinated Objects}|}{|\text{All Objects}|}, \\
   \text{CHAIRi} &= \frac{|\text{Hallucinated Captions}|}{|\text{All Captions}|}
\end{aligned}
\end{equation}

\paragraph{POPE.} POPE is a widely adopted benchmark for evaluating object hallucinations by prompting LVLMs to identify whether a specific object is present in the image. It comprises three distinct datasets: \emph{MSCOCO}~\cite{coco}, \emph{A-OKVQA}~\cite{aokvqa}, and \emph{GQA}~\cite{gqa}.
Each dataset uses three different negative sampling settings: \emph{Random}, \emph{Popular}, and \emph{Adversarial}.
Each subset includes 3,000 questions and 500 images. Accuracy and F1 score are used as the primary evaluation metrics.

\paragraph{AMBER.} AMBER combines generative and discriminative tasks, and is evaluated on a curated set of 1,004 images. In addition to image captioning, it includes 14,216 questions designed to assess hallucinations in object, attribute, and relation recognition.
AMBER contians mulitple metrics: \emph{CHAIR}, \emph{Cover}, \emph{Hal}, \emph{Cog}. 
It provides an annotated objects list $A_{obj}={obj_1^A, obj_2^A, \cdots, obj_n^A}$, and the generated objects are labeled as $R'_{obj}$.
Each metric is calculated as follows:
\begin{equation}
\begin{aligned}
   \text{CHAIR} &= 1 - \frac{\mathrm{len}(R'_{obj} \cap A_{obj})}{\mathrm{len}(R'_{obj})},\\
   \text{Cover} &= \frac{\mathrm{len}(R'_{obj} \cap A_{obj})}{\mathrm{len}(A_{obj})}, \\
   \text{Hal}   &= \frac{\{ \text{CHAIR} > 0 \}}{\{\text{All Caps}\}}, \\
   \text{Cog}   &= \frac{\mathrm{len}(R'_{obj} \cap H_{obj})}{\mathrm{len}(R'_{obj})},
\end{aligned}
\end{equation}
where $H_{obj}$ denotes the set of hallucinated target objects generated by the LVLMs, and \emph{All Caps} refers to all generated captions.

\subsection{Baselines}
We select three existing visual attention-based dehallucination methods for comparative analysis:
\footnote{Note: The same parameters may have different meanings across methods. For specific details, please refer to the original papers.}

\begin{itemize}
    \item \textbf{PAI}~\cite{PAI}: This method directly amplifies the model's attention weights on visual tokens and further enhances visual features using contrastive decoding. To facilitate a fair comparison of visual attention optimization, we denote the variants with and without contrastive decoding as PAI$_\mathrm{CD}$ and PAI, respectively. We set the visual attention augmentation coefficient $\alpha$ to 0.5 and apply visual attention augmentation starting from the layer specified by our method. PAI$_\mathrm{CD}$ has a hyperparameter $\gamma=1.1$ and employs adaptive plausibility constraints with $\beta=0.1$.
    \item \textbf{VAF}~\cite{VAF}: Similar to PAI, VAF directly scales up visual attention; however, it also suppresses attention on instruction tokens.  We set $\beta=0.1$, $\alpha=0.15$, and activate this method from layer 9 to layer 14 (counting from 0).
    \item \textbf{TARAC}~\cite{TARAC}: This method quantifies the importance of visual tokens based on attention distributions from previous generation steps and utilizes this information to strengthen the model's focus on relevant visual tokens at the current step. We set $\beta=0.5$, $\alpha=0.5$, and activate this method from layer 9 to layer 15 (counting from 0).
\end{itemize}

\begin{figure*}
    \centering
    \includegraphics[width=0.8\linewidth]{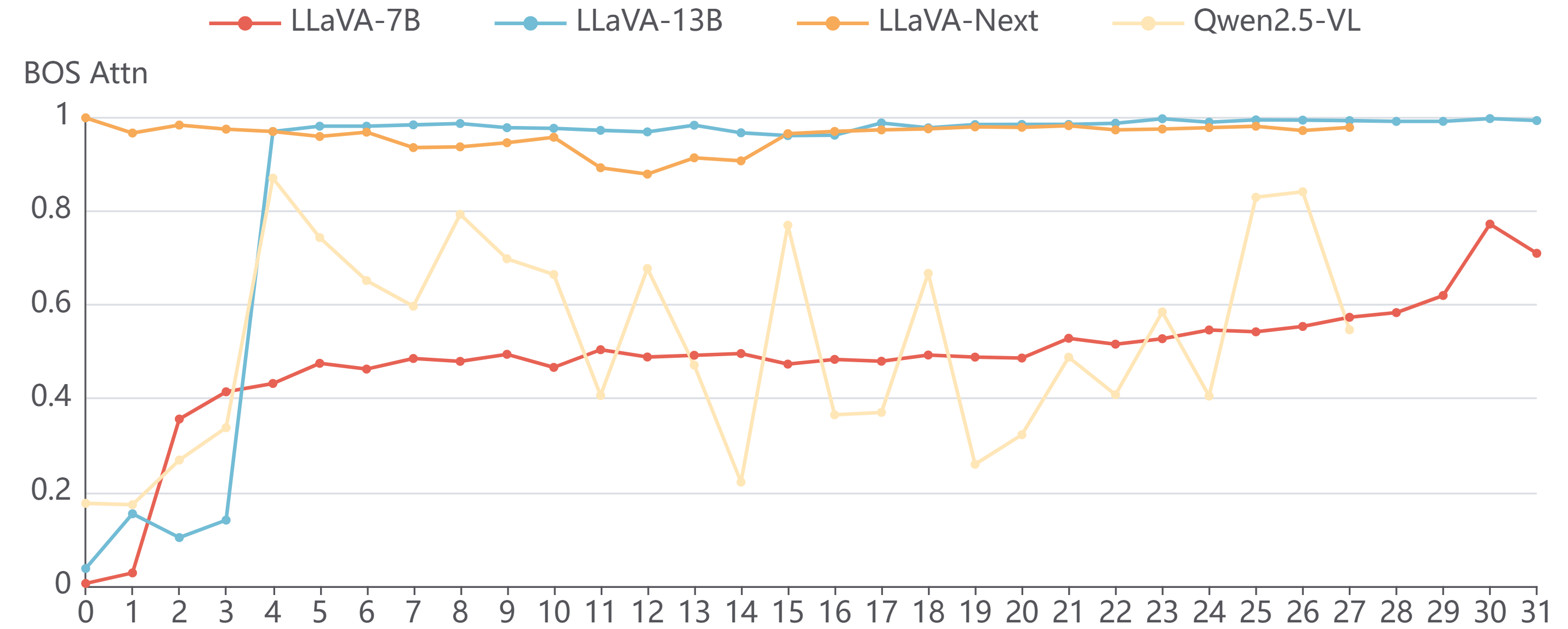}
    \caption{The model's attention to the BOS token in each layer.}
    \label{fig:bos}
\end{figure*}

\subsection{Implementation Details}
Following the established practice of PAI, we determine the starting layer for VGA based on the \emph{BOS} attention, which is discussed in \Cref{sec:start layer}. Specifically, the starting layers are set as follows (with layers numbered from 0): layer 2 for LLaVA-1.5, layer 0 for LLaVA-Next, and layer 4 for Qwen2.5-VL. We apply VGA up to an intermediate layer of the model: specifically, layer 24 for LLaVA-1.5-13B and layer 16 for all other models.
We adopt the default settings of $\beta=0.2$ and $\lambda=0.02$.
For the larger model (LLaVA-1.5-13B) and visually simpler tasks (POPE), we increase $\beta$ to $0.25$ to apply stronger vision guidance.
Stanza~\cite{stanza} is utilized to extract objects mentioned in the question.
We employ greedy decoding for next token prediction and set the maximum generation length to 512. All experiments are conducted using a single NVIDIA A800-40G GPU.

\begin{figure}
    \centering
    \includegraphics[width=0.8\linewidth]{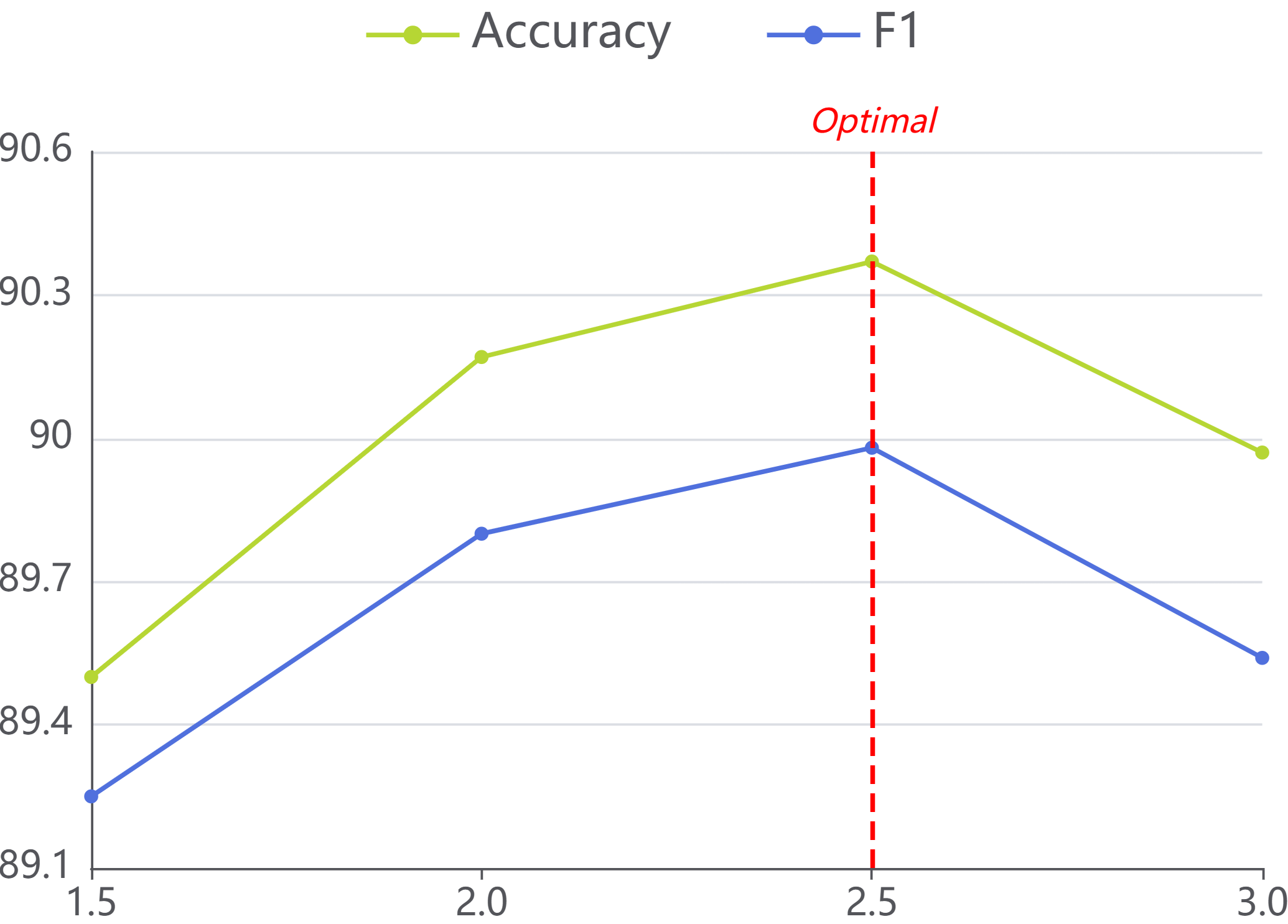}
    \caption{Effectiveness of $\beta$ in MSCOCO's Random set}
    \label{fig:pope_guid}
\end{figure}

\section{Discussion of Main Results}\label{app:main results}
\paragraph{POPE.}
The experimental results on the POPE benchmark are summarized in \Cref{tab:pope}. Our proposed method, VGA, achieves significant improvements, clearly demonstrating its advantage in mitigating existence hallucinations. Moreover, VGA consistently delivers positive gains across all tested MLLMs, highlighting its strong generalizability.
While the baselines enhance visual understanding by strengthening the model's attention to visual content, they often lack precise localization of key objects, which leads to suboptimal performance. The superior performance of VGA indicates that providing explicit visual guidance effectively enhances the model's ability to perceive and discriminate objects within the image.

\paragraph{CHAIR.}
The experimental results on the CHAIR benchmark are presented in \Cref{tab:chair}. Our method achieves the overall lowest hallucination rate across all evaluated MLLMs, while maintaining no significant drop in $\text{F1}$ score.
This result underscores VGA's ability to strike a better balance between generating a richer output and ensuring generation accuracy.
Although $\text{PAI}$ enhanced by contrastive decoding (PAI$_{\text{CD}}$) further refines visual information in the logits and achieves strong performance, contrastive decoding requires two forward passes, which effectively doubles the inference cost. In stark contrast, VGA achieves superior performance with only a single forward pass, demonstrating both higher efficiency and effectiveness.

\paragraph{AMBER.}
AMBER is a comprehensive hallucination benchmark that enables a more thorough evaluation of model hallucinations.
The results, shown in \Cref{tab:amber} and \Cref{tab:amber_more}, demonstrate that VGA still achieves significant dehallucination performance, further validating the importance of precise visual guidance.
Methods like TARAC guide visual attention based on historical attention distributions; however, the localization capability derived solely from visual attention is inherently limited. In contrast, VGA leverages VSC to achieve more accurate visual grounding without relying on external tools, thus enabling a more effective suppression of hallucinations.

\begin{table*}
    \centering
    \caption{
    Results on AMBER with LLaVA-13B. The AMBER metric is calculated as $ (1-\text{CHAIR}+\text{F1})/2 $.
    }\label{tab:amber_more}
    \begin{tabular}{ll|cccc|cccc|c}
       \toprule
      MLLM & Method & CHAIR $\downarrow$ & Cover $\uparrow$ &  Hal $\downarrow$& Cog $\downarrow$ & Acc. $\uparrow$&  Prec. $\uparrow$ & Rec. $\uparrow$ & F1 $\uparrow$ &  AMBER $\uparrow$ \\
       \midrule
     \multirow{6}{*}{LLaVA-13B}
     &   Vanilla &    6.8 & 51.3 & 31.1 & 3.4 & 71.5 & \textbf{96.0} & 59.5 & 73.5 & 33.85 \\
     &  PAI &   5.4 & \underline{51.6} & \underline{27.3} & 2.1 & 74.2 & \underline{95.6} & 64.0 & 76.7 & 36.15   \\
     &  PAI$_\mathrm{CD}$ &   \underline{5.2} & \textbf{52.0} & 28.3 & \textbf{1.9} & \underline{75.0} & 95.3 & \underline{65.5} & \underline{77.6} & \underline{36.70}   \\
      &  VAF &    6.9 & 51.2 & 32.0 & 3.6 & 69.5 & \underline{95.6} & 56.7 & 71.2 & 32.65   \\
     &  TARAC &  5.6 & 49.5 & 25.1 & 2.4  & 73.0 & \textbf{96.0} & 61.9 & 75.3 & 35.35   \\
      \rowcolor{gray!20}
      &  VGA &    \textbf{4.6} & 49.9 & \textbf{23.6} & \underline{2.0} & \textbf{78.5} & 95.0 & \textbf{71.3} & \textbf{81.5} & \textbf{38.95}   \\
       \bottomrule
    \end{tabular}
\end{table*}

\section{The Starting Layer of VGA}
\label{sec:start layer}
In self-attention mechanisms, a ``sink'' phenomenon~\cite{VAR,SEVI} exists—where the model abnormally focuses its attention on a few individual tokens. PAI~\cite{PAI} observes that the BOS token carries very limited information yet consistently receives disproportionately high attention. Consequently, when the BOS token receives excessively high attention from the model, it serves as an appropriate trigger to enhance the model's visual attention. Following the approach of PAI, we analyze the BOS attention patterns across different models to determine the optimal starting layer for VGA.
We apply max pooling across all attention heads.
The analysis results on the image captioning task are shown in \Cref{fig:bos}.
Based on our observations, we set the starting layer of VGA for different MLLMs as follows: layer 2 for LLaVA-1.5-7B, layer 4 for LLaVA-1.5-13B, layer 0 for LLaVA-Next, and layer 4 for Qwen2.5-VL.

\begin{table}
    \centering
        \caption{Ablation study on CHAIR.
        }\label{tab:ablation}
    \begin{tabular}{l|ccc}
    \toprule
       Setting &  Cs & Ci   &  F1\\
         \midrule
        VGA &  \textbf{29.8} & \textbf{8.2}  & \textbf{76.3}    \\
      \emph{w/o} Head Balancing & 34.6 & 9.4 & \textbf{76.3}  \\
       \emph{w/o} Early Termination &  32.6 & 8.8 & 75.4 \\
      \bottomrule
    \end{tabular}
\end{table}

\section{Ablation Study}
Considering the functional diversity across attention heads, VGA employs head balancing by applying stronger guidance to heads with weaker visual features and lighter guidance to those that already exhibit strong visual functionality. Additionally, since visual understanding in MLLMs primarily occurs in the early to middle layers, we terminate visual attention guidance at an intermediate layer of the model. We conduct ablation studies on both of these design choices, and the results are presented in \Cref{tab:ablation}.
Head balancing preserves the model's original visual capabilities while incorporating additional vision-guidance, thereby achieving improved performance. Meanwhile, the early exit mechanism prevents visual guidance from being applied during non-visual understanding stages, maintaining consistency with the model's inherent behavioral characteristics.
In short, both the head balancing and early termination modules contribute positively to performance.


\section{Hyperparameters}

\begin{table}
    \centering
        \caption{Effectiveness of $\beta$ and $\lambda$ in CHAIR's validation set.
        }\label{tab:chair_grid}
    \begin{tabular}{l|cc}
    \toprule
       Setting & Cs &  F1 \\
         \midrule
         Vanilla & 49.6 & 76.5 \\
         \hline
         $\beta=0.15, \lambda=0.01$ & 41.6 & 77.5 \\
         $\beta=0.15, \lambda=0.02$ & 43.4 & 76.8 \\
         $\beta=0.15, \lambda=0.04$ & 43.6 & 77.1 \\
         \hline
         $\beta=0.20, \lambda=0.01$ & 26.2 & 76.4 \\
      \rowcolor{gray!20}
         $\beta=0.20, \lambda=0.02$ & 32.8 & 77.7 \\
         $\beta=0.20, \lambda=0.04$ & 38.6 & 77.9 \\
         \hline
         $\beta=0.25, \lambda=0.01$ & 10.0 & 68.7 \\
         $\beta=0.25, \lambda=0.02$ & 11.8 & 69.1 \\
         $\beta=0.25, \lambda=0.04$ & 19.8 & 71.8 \\
      \bottomrule
    \end{tabular}
\end{table}

$\beta$ controls the strength of visual guidance, while $\lambda$ controls the penalty strength on already-generated content in image captioning tasks. We perform a grid search for $\beta$ on the POPE benchmark, as shown in \Cref{fig:pope_guid}. Additionally, we conduct a grid search over both $\beta$ and $\lambda$ on the CHAIR benchmark, with results presented in \Cref{tab:chair_grid}.
We observe that for tasks like POPE, which require focusing on a single visual region, stronger visual guidance is beneficial. In contrast, for image captioning tasks that demand broader attention across multiple visual regions, overly strong guidance can restrict the model's capacity. Therefore, we set $\beta = 0.25$ on the POPE benchmark and $\beta = 0.2$ on other benchmarks. Furthermore, for image captioning, we select $\lambda = 0.02$ based on a balanced consideration of hallucination rate and F1 score.

\begin{table}[]
\setlength{\tabcolsep}{2mm}
\renewcommand{\arraystretch}{0.9}
    \centering
    \small{
    \caption{\label{tab:mme}
    \emph{Att.} and \emph{Rel.} denote the F1 scores on AMBER's attribute and relation tasks, respectively.
    \emph{MME} denote perception tasks in the MME benchmark.
    }
    
    \begin{tabular}{llccc}
    \toprule
    \textbf{MLLM} &\textbf{Method} & \textbf{Att.} & \textbf{Rel.} & \textbf{MME} \\
    \midrule
    \multirow{3}{*}{LLaVA-1.5-7B} 
    & Vanilla  & 64.4 & 68.5  & 1456.5 \\
   &    TARAC  & 63.5 & 68.4  & 1462.1  \\
    \rowcolor{gray!20}
       \cellcolor{white}
     &  VGA  & \textbf{65.7} & \textbf{73.9}  &  \textbf{1465.7}  \\
       \midrule
    \multirow{3}{*}{Qwen2.5-VL-7B}
    & Vanilla  & 84.4 & 75.8 & 1691.1 \\
    &   TARAC  &  84.8 & 76.2 & 1713.1  \\
       \rowcolor{gray!20}
       \cellcolor{white}
    &  VGA  & \textbf{86.4} & \textbf{76.3} & \textbf{1719.7}  \\
       \bottomrule
    \end{tabular}
    }

\end{table}

\section{Beyond Object Hallucination.}
The model's comprehension of auxiliary information, such as relations and attributes, fundamentally relies on its understanding of objects; thus, object grounding remains critical. Furthermore, for tasks lacking explicit objects (\emph{e.g.}, image captioning), we employ the VSS mechanism to establish informative grounding (see Sec. 3.2). Additional results presented in Table~\ref{tab:mme} indicate that VGA maintains advantages in mitigating relation- and attribute-based hallucinations. We further validate VGA on general VQA benchmarks. Results on MME \cite{MME} demonstrate that VGA performs effectively on object-agnostic tasks, confirming that these performance gains stem from enhanced visual perception via vision guidance.

\section{First Token Approximation.}
We approximate the visual semantic confidence of an object using the first token of its tokenized representation. Under a single-step prediction setting, utilizing the first token aligns with the autoregressive paradigm; conversely, the probability of any non-first token (without the ``\_'' prefix) is extremely low (always less than $1e-4$), rendering their distribution across visual tokens meaningless. We also experimented with aggregating the probabilities of all tokens ($\log c(O) \approx \sum_i \log c(o_i)$); however, this yielded negligible performance differences. Consequently, we adopt the simplest approach.

\end{document}